\documentclass[letterpaper, 10 pt, conference]{ieeeconf}  

\IEEEoverridecommandlockouts                              

\usepackage{subcaption}
\usepackage[pdftex]{graphicx} 
\usepackage{booktabs}
\usepackage{multirow}
\usepackage{amsmath,amssymb,amsfonts,bbm,mathtools,bm}
\usepackage{xcolor}
\usepackage[font=footnotesize]{caption} 
\usepackage{flushend}

\title{\LARGE \bf
Robust Surgical Tool Tracking with Pixel-based Probabilities for Projected Geometric Primitives
}

\author{Christopher D'Ambrosia$^{1}$, Florian Richter$^{2}$, Zih-Yun Chiu$^{2}$, Nikhil Shinde$^{2}$,\\ Fei Liu$^{2}$, Henrik I. Christensen$^{3}$ and Michael C. Yip$^{2}$
\thanks{$^{1}$Christopher D'Ambrosia is with the College of Physicians and Surgeons, Columbia University and the Department of Computer Science and Engineering, University of California San Diego, La Jolla, California 92093, U.S.A. (cdambros@ucsd.edu)}%
\thanks{$^{2}$Florian Richter, Zih-Yun Chiu, Nikhil Shinde, Fei Liu and Michael C. Yip are with the Deparment of Electrical and Computer Engineering, University of California San Diego, La Jolla, California 92093, U.S.A. (frichter@ucsd.edu; zchiu@ucsd.edu; nshinde@ucsd.edu; f4liu@ucsd.edu; yip@ucsd.edu)}%
\thanks{$^{3}$Henrik I. Christensen is with the Department of Computer Science and Engineering, University of California San Diego, La Jolla, California 92093, U.S.A. (hichristensen@.ucsd.edu)}%
}

\begin{document}

\maketitle
\thispagestyle{empty}
\pagestyle{empty}

\begin{abstract}
Controlling robotic manipulators via visual feedback requires a known coordinate frame transformation between the robot and the camera. Uncertainties in mechanical systems as well as camera calibration create errors in this coordinate frame transformation. These errors result in poor localization of robotic manipulators and create a significant challenge for applications that rely on precise interactions between manipulators and the environment. In this work, we estimate the camera-to-base transform and joint angle measurement errors for surgical robotic tools using an image based insertion-shaft detection algorithm and probabilistic models. We apply our proposed approach in both a structured environment as well as an unstructured environment and measure to demonstrate the efficacy of our methods.
\end{abstract}

\section{INTRODUCTION}

Surgical automation has the potential to increase access to and quality of surgical care \cite{yip2019robot}. For this reason, automation of surgical subtasks remains an active area of research \cite{wilcox2022learning}. Prior work includes autonomous surgical cutting \cite{murali2015learning, thananjeyan2017multilateral}, hemostasis \cite{richter2021autonomous}, peg transfer \cite{hwang2020applying}, and suturing \cite{chiu2021bimanual, chiu2022markerless}.
Surgical automation, however, requires robust and accurate end effector localization \cite{lu2023markerless}. Localization typically involves camera or image-based feedback \cite{lin2022semantic}. However, uncertainty in the camera-to-robot transforms \cite{richter2021robotic} as well as imprecision due to mechanical factors including cable-stretch and hysteresis \cite{hwang2020efficiently} limits image-based localization accuracy.



\begin{figure}
    \centering
    \vspace{2mm}
    \includegraphics[width=0.48\textwidth, trim={0cm 0 4cm 0},clip]{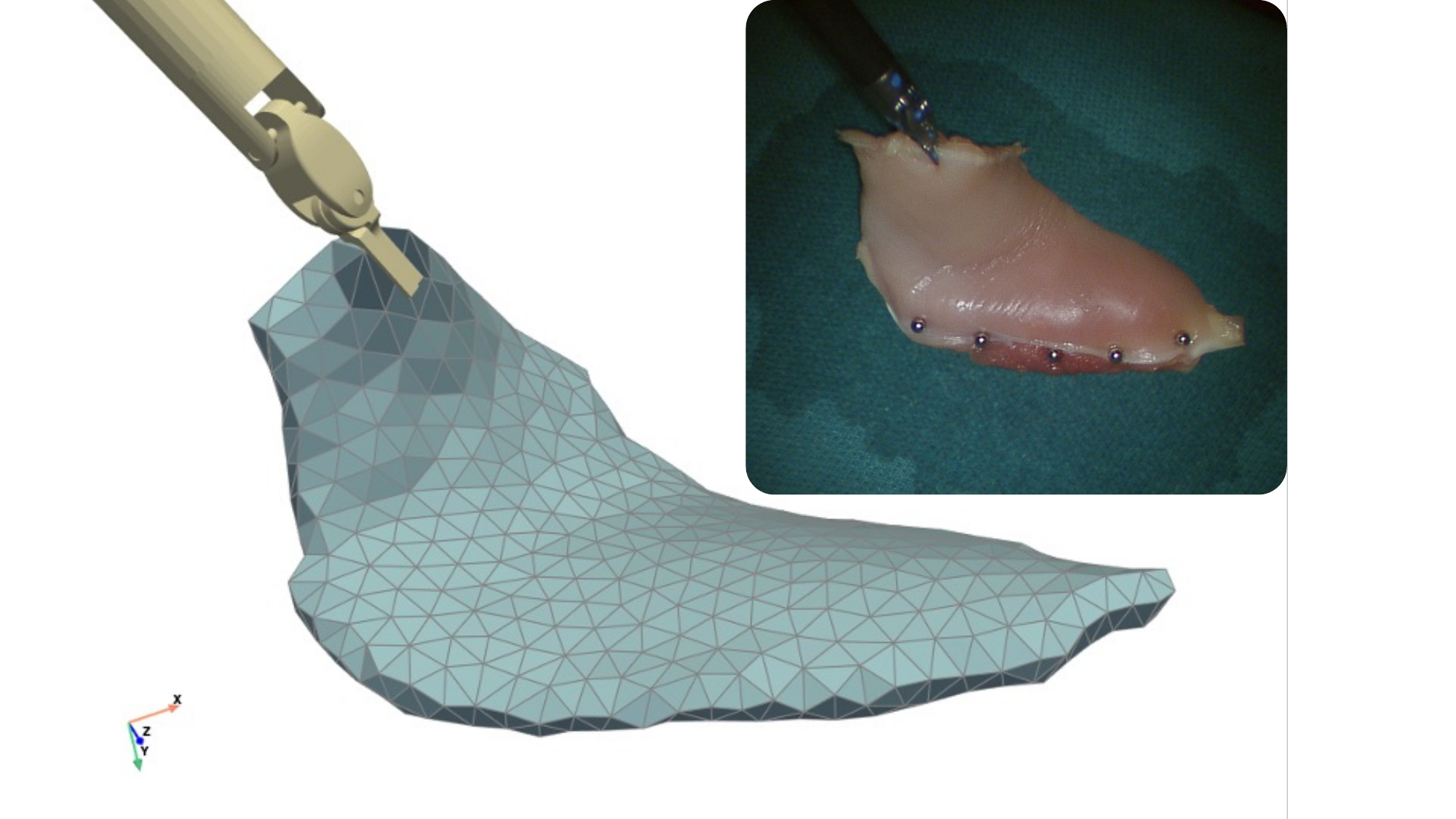}
    \caption{The insertion-shaft on surgical robotic tools is an excellent feature for localization. In this work, we present a novel approach for detecting the insertion-shaft and incorporating it into Bayesian filtering to probabilistically localize and track surgical robotic tools. We stress test our method in a challenging deformable tissue dataset where a surgical robotic tool is in the back of the scene with low light and commanded to deform tissue.
    The tissue is registered to a simulated scene using a separate approach in the camera frame \cite{liu2021real} and notice how well aligned the grasp between the localized surgical robotic tool and the tissue is in the 3D rendering.
    }    \label{fig:cover_figure}
\end{figure}

\subsection{Related Works} 

Prior work in estimating the base-to-camera transform includes solving homogeneous linear systems based on multiple images that capture markers affixed to robotic links \cite{fassi2005hand, park1994robot}, Solve-PnP approaches using detected markers on robotic manipulators \cite{lepetit2009ep} or marker-less keypoint detection \cite{lambrecht2019towards, lambrecht2019robust, lee2020camera, lu2020robust}.
%
In the space of surgical robotics, previous works have calibrated remote center of motion (RCM) robotic systems \cite{zhong2020hand} and kinematic remote center coordinate systems (KCS) that accounts for error in the RCM camera frame-to-base transform \cite{zhao2015efficient}.
Markers, learned features, silhouette matching, and online template matching from images have been used for KCS tracking \cite{li2020super, reiter2012feature, reiter2014appearance, lu2021super}.

To estimate joint angle measurement errors in the context of cable-driven surgical robotics, prior work has applied data-driven approaches, real-time inverse kinematics, and physical modeling of cable transmission friction and hysteresis \cite{pastor2013learning, wang2013online}.
Learning-based approaches for estimating cable stretch and Gaussian processes for measurement error compensation have also been used \cite{hwang2020efficiently, peng2020real, mahler2014learning}.
Image-based methods have used unscented Kalman filters and deep learning tools to accurately estimate cable-driven joint angles \cite{haghighipanah2016unscented, 8460583}.

Estimating both camera-to-base transform and joint angle offsets simultaneously using joint calibration techniques have been proposed \cite{pradeep2014calibrating, le2009joint}.
These methods, however, can not account for dynamic uncertainties such as non-constant joint angle errors. Methods to achieve real-time estimation of dynamic joint angle errors have used iterative closest point matching based on 3D point clouds and Kalman filters \cite{krainin2010manipulator}.
Probabilistic approaches using observation models parameterized by physical parameters have also been explored \cite{cifuentes2016probabilistic}.
While these works prioritize integrating additional sensors such as depth cameras into real-time estimation, we rely solely on endoscopic images which is available in surgical robotic scenes.
Furthermore, surgical robotics has the unique challenge of a partially visible kinematic chain which creates a one-to-many mapping between the detected images and the parameter set describing the camera-to-base transform and joint angle errors \cite{richter2021robotic}.

\subsection{Contribution}

To address the challenge of estimating the camera-to-base transform and joint angle errors, our prior work introduced the concept of lumped error.
Lumped error accounts for uncertainty in the camera-to-base transform, camera calibration, and joint transforms in a forward-kinematic model for partially visible kinematic chains \cite{richter2021robotic}.
Estimating this lumped error has enabled more accurate, probabilistic tool tracking for surgical systems \cite{chiu2022real} and utilized for autonomous blood suction \cite{richter2021autonomous} and suture needle manipulation \cite{chiu2021bimanual}.

One key observation we made is the importance of the insertion-shaft as an observation for estimating the lumped error.
The insertion-shaft is how the surgical robotic manipulator is inserted into the surgical scene through port entries and a standard feature for any laparoscopic tool hence using it for estimation will help constrain the RCM location with respect to the camera frame \cite{zhong2020hand, lu2022unified}.
Furthermore, the insertion-shaft can be modelled as geometric primitive, a cylinder, which has an analytical camera projection equation \cite{project_cylinder}.
The insertion-shaft was incorporated into our previous work in a probabilistic  fashion to estimate the lumped error via a detection algorithm and Bayesian filtering.
While using the insertion-shaft as an observation was crucial for surgical robotic tool tracking, there are two limitations with our previous implementation:
1) detection of the insertion-shaft parameters from the endoscopic data was inconsistent in unstructured scenes and
2) the probabilistic observation model to update the lumped error belief was limited to a line parameterization space.
To address these challenges we contribute a laparoscopic tool insertion-shaft detection approach based on a Deep Neural Network (DNN) with multiple observation models for Bayesian filtering.
The approach is integrated into a Particle Filter and a comparative study is done to evaluate the different observation models on two datasets collected on the da Vinci Research Kit (dVRK) \cite{kazanzides2014open}.

\section{Background}

Any joint-link on a surgical robot can be localized in the camera frame by combining the robots forward kinematics and the base-to-camera transform.
Written explicitly, the homogeneous transform matrix describing the $j$-th joint-link in the camera frame is
\begin{equation}
    \label{eq:fk_ideal}
	^{c}\textbf{T}_j = ^{c}\textbf{T}_{b} \prod^{j}_{i = 1} {^{i-1}\textbf{T}_{i}}(q_{i})
\end{equation}
where $^{c}\textbf{T}_{b} \in SE(3)$ is the base-to-camera transform and ${^{i-1}\textbf{T}_{i}}(q_{i}) \in SE(3)$ is the $i$-th joint transform with joint angle $q_{i}$.
In an ideal case, the joint transforms are provided by the robot manufacturer and the joint angles are measured by encoders hence only the base-to-camera transform needs to be solved for.


Due to the unique workspace challenges faced in surgical environments (e.g. narrow corridors), surgical robotic systems often rely on cable-drive designs without encoders at the joint-link location such as the dVRK \cite{kazanzides2014open} and RAVEN robotic systems \cite{lum2009raven, hannaford2012raven}.
Instead joint angles are read at the motor which leads to inaccurate joint readings from mechanical phenomena such as cable stretch, hysteresis and cable stretch \cite{haghighipanah2016unscented, hwang2020efficiently}. 
Furthermore, the kinematic chains of surgical robots are only partially visible in the camera frame.
For example, the endo-wrist and insertion-shaft from da Vinci \textregistered surgical platform are often the only joint-links visible in the endoscopic frame.

Localizing the surgical robot in the camera frame from endoscopic images by estimating the hand-eye transform and joint angle errors is not feasible since the parameters are not identifiable from the camera frame due to the partially visible robotic chain \cite{richter2021robotic}. 
In our previous work, we derive a smaller parameter set that is identifiable from the camera frame and enables complete localization of the surgical robotic links visible in the camera frame.
The forward kinematics in (\ref{eq:fk_ideal}) can be written with the new parameter set as
\begin{equation}
    \label{eq:fk}
	^{c}\textbf{T}_j = \textbf{E} \prod^{j}_{i = 1} {^{i-1}\textbf{T}_{i}}(\tilde{q}_{i})
\end{equation}
where $\textbf{E} \in SE(3)$ is the lumped error transform and $\tilde{q}_{i}$ are the noisy joint angle readings.
The lumped error transform captures both the camera-to-base transform and errors in joint angle readings.
Please see \cite{richter2021robotic} for a complete derivation.

To estimate the lumped error, features are detected on the surgical robotic tool.
The scope of this work focuses on using the insertion-shaft as a feature which can be modelled as a cylinder with  the following parameters: a position on the center-line, $\textbf{p}^j_0 \in \mathbb{R}^3$, direction vector of the centerline, $\textbf{d}^j \in \mathbb{R}^3$, and radius $r \in \mathbb{R}^{+}$.
Through the kinematics equation in (\ref{eq:fk}), the insert-shaft cylinder on the $j$-th joint can be transformed to the camera frame: $\textbf{p}^c_0 = {}^c\textbf{T}_j \textbf{p}^j_0$ and $\textbf{d}^c = {}^c\textbf{T}_j \textbf{d}^j$.
Finally, a point, $\textbf{p} \in \mathbb{R}^3$, is defined to be on the insertion-shaft in the camera frame if it satisfies the following equations
\begin{equation}
\label{eq:insertion_shaft_cylinder_cross_section_intersection}
        \begin{cases}
        (\mathbf{p} -  \mathbf{p}^c_a)^\top (\mathbf{p} -  \mathbf{p}^c_a) - r^2 = 0 \\
        (\mathbf{d}^c)^\top (\mathbf{p} -  \mathbf{p}^c_a ) = 0
    \end{cases}
\end{equation}
where $\mathbf{p}^c_a$ is a point on the center-line
\begin{equation}\label{eq:insertion_shaft_cylinder_center_line}
    \mathbf{p}^c_a = \mathbf{p}_0^c+ \lambda \mathbf{d}^c
\end{equation}
and $\lambda \in \mathbb{R}$.
Chaumette showed that projecting (\ref{eq:insertion_shaft_cylinder_cross_section_intersection}) with the camera pin-hole model results in two line segments that represent the two edges of the projected cylinder \cite{project_cylinder},
\begin{equation}
    \label{eq:projected_cylinder_lines}
    a_1 X + b_1 Y + c_1 = 0 \qquad a_2 X + b_2 Y + c_2 = 0
\end{equation}
where $(X,Y)$ are pixel coordinates on a unit camera (i.e. $(X, Y) = (\frac{u - c_u}{f_x}, \frac{v - c_v}{f_y})$ where $(c_u, c_v), (f_x, f_y)$ are the principle point and focal length of a non-unit camera), 
\begin{equation}
    \begin{split}
        A_{1,2} = \frac{r \left( x^c_0 - a^c (\textbf{p}^c_0)^\top \textbf{d}^c \right)}{\sqrt{(\textbf{p}^c_0)^\top \textbf{p}^c_0 - (\textbf{p}^c_0)^\top \textbf{d}^c - r^2}} \pm ( c^c y^c_0 - b^c z^c_0 )\\
        B_{1,2} = \frac{r \left( y^c_0 - b^c (\textbf{p}^c_0)^\top \textbf{d}^c \right)}{\sqrt{(\textbf{p}^c_0)^\top \textbf{p}^c_0 - (\textbf{p}^c_0)^\top \textbf{d}^c - r^2}} \pm ( a^c z^c_0 - c^c x^c_0 )\\
        C_{1,2} = \frac{r \left( z^c_0 - c^c (\textbf{p}^c_0)^\top \textbf{d}^c \right)}{\sqrt{(\textbf{p}^c_0)^\top \textbf{p}^c_0 - (\textbf{p}^c_0)^\top \textbf{d}^c - r^2}} \pm ( b^c x^c_0 - a^c y^c_0 )\\
    \end{split}
\end{equation}
and $\textbf{p}^c_0 = \begin{bmatrix} x^c_0 & y^c_0 & z^c_0 \end{bmatrix}^\top$, $\textbf{d}^c = \begin{bmatrix} a^c & b^c & c^c \end{bmatrix}^\top$.
By combining the forward kinematic model in (\ref{eq:fk}) with a the insertion-shaft projection (\ref{eq:projected_cylinder_lines}), estimation methods can be used to find the lumped error, $\textbf{E}$, from insertion-shaft feature detections hence localizing the surgical robotic tools in the camera frame.


\section{Methods}

To estimate the lumped error, a particle filter tracks its probability distribution temporally using observations from the endoscopic camera.
The lumped error is parameterized with a translation and axis-angle vector, $\textbf{E}(\textbf{b}_t, \textbf{w}_t)$ where $\textbf{b}_t, \textbf{w}_t \in \mathbb{R}^3$.
In the coming sub-sections, the motion model, insertion-shaft line detection algorithm, and observation model are described to fully define the necessary components for a Bayesian Filter to track the lumped error hence localizing the surgical robotic tool in the camera frame.

\subsection{Motion Model}

The lumped error's motion model is defined with additive Gaussian noise due to its ability to generalize over a large number of random processes.
Written explicitly, the motion model is
\begin{equation}
    \begin{bmatrix} \textbf{b}_{t+1}, \textbf{w}_{t+1} \end{bmatrix}^\top = \mathcal{N} \left( \begin{bmatrix}
    \mathbf{b}_{t}, \mathbf{w}_{t}  \end{bmatrix}^\top,  \mathbf{\Sigma}_{\mathbf{b},\mathbf{w}, t+1} \right)
\end{equation}
where $\mathbf{\Sigma}_{\mathbf{b}, \mathbf{w},t+1} \in \mathbb{R}^{6 \times 6}$ is a covariance matrix.






\begin{figure}
    \vspace{2mm}
    \centering 
    \begin{subfigure}{0.49\columnwidth}
    \includegraphics[width=\columnwidth, trim={0cm 0cm 6.75cm 0},clip]{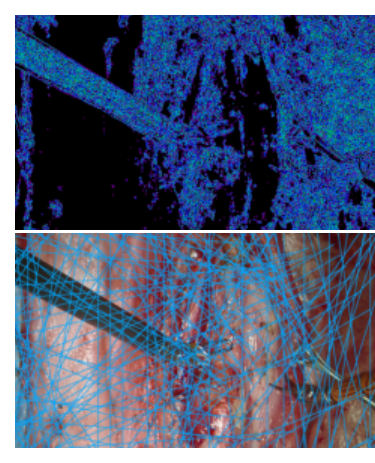}
    \end{subfigure}
    \begin{subfigure}{0.49\columnwidth}
    \includegraphics[width=\columnwidth, trim={6.75cm 0 0 0},clip]{shaft_detection.pdf}
    \end{subfigure}
    \caption{From left to right, the columns show our previous approach, based on Canny Edge detection \cite{richter2021robotic}, and proposed approach, based on SOLD2 \cite{pautrat2021sold2}, for detecting the insertion shaft when deployed in a live surgery \cite{richter2021bench}. The top row shows a heatmap of the pixels potentially associated with a line segment and the bottom row shows all the detected line segments in light blue from both approaches
    Additionally, the purple line segments for the SOLD2 approach highlight the detected lines associated with the insertion shaft of the surgical tool which was not possible with the previous approach.
    }
    \label{fig:shaft_detection}
\end{figure}

\subsection{Insertion-Shaft Line Detection}



In our previous work, we manually tuned a canny edge detection algorithm \cite{canny1986computational} followed by a hough transform to detect lines \cite{ballard1981generalizing}.
From there, a greedy association strategy is employed to find which detected lines are associated with the edges of the insertion-shaft.
In this work, we use SOLD2, a self-supervised occlusion-aware line description and detection DNN \cite{pautrat2021sold2}, for more reliable, robust insertion-shaft line detection in unstructured environments.
The SOLD2 model at inference is provided a reference image of the surgical robotic tool and outputs the detected insertion-shaft lines by finding which of the detected lines are associated with the insertion-shaft from the reference image.
SOLD2 also provides a heatmap of all the detected pixels associated with the detected lines.
Written explicitly, SOLD2 outputs
\begin{equation}
    \textbf{e}^a_{1,2}, \textbf{e}^b_{1,2}, \textbf{H} = f(\mathcal{I}; \mathcal{I}_{ref})
\end{equation}
where $\textbf{e}^a_{1,2}, \textbf{e}^b_{1,2} \in \mathbb{R}^2$ are the endpoints of the detected insertion-shaft line segments, $\textbf{H} \in \mathbb{I}^{H \times W}$ is the heatmap, $\mathcal{I} \in \mathbb{I}^{H \times W \times 3}$ is the input image, and $\mathcal{I}_{ref} \in \mathbb{I}^{H \times W \times 3}$ is the reference image.
An example insertion-shaft detection is shown in Fig. \ref{fig:shaft_detection}.


SOLD2 relies on similarity-based image features to evaluate candidate line endpoints.
Surgical endoscope images, however, may have barrel distortion and lens vignetting \cite{smith1992correction} which significantly increases image feature variance between the center and periphery of the image.
Lines with one endpoint located at the center of the image and one endpoint located at the periphery of the image (e.g. insertion-shaft edges) are often falsely eliminated due to the distortion and artifact-based dissimilarities in the neighborhoods of proposed endpoints.
We augment our surgical endoscope images through cropping and downsampling to reduce the number of false negative endpoint detections.
While any line detection and association algorithm could be used in our tool-tracking pipeline, we choose to use SOLD2 rather than other learning-based line detection algorithms \cite{xie2015holistically} because SOLD2 provides line associations across video frames which are a necessary component for our downstream filtering of false line detections.  


\subsection{Observation Model}

From the insertion-shaft line detection, we present four different observation models and compare them experimentally.
The first two are similar to our previous work where the uncertainty is modelled on line parameters, distance to the origin and slope angle.
Meanwhile the second two observation models work with pixels directly by deriving a random variable that describes how well the pixel fits on to the projected insertion-shaft.


\subsubsection{Endpoint Intensities to Polar}
\label{sec:endpoint_intensities_to_polar}

SOLD2 architecture directly outputs endpoint candidates and then completes the line segments though the heatmap \cite{pautrat2021sold2}.
Similarly, the first observation model uses the detected endpoints, $\textbf{e}^a_{1,2}, \textbf{e}^b_{1,2}$, to derive detected line parameters, $\rho^e_{1,2}, \theta^e_{1,2}$, which represents the distance to origin and slope angle.
To increase the robustness, we derive the detected line parameters from a set of points near the end-points that have a high heat-map value hence should be part of a line.
Written explicitly, the endpoint set is
\begin{equation}
    \label{eq:end_point_set}
	\mathcal{E}_e(\textbf{e}) = \{ \textbf{p} \hspace{1mm} \big| \hspace{1mm} || \textbf{p} - \textbf{e} || \leq \alpha_e \hspace{1mm} \cap \hspace{1mm} \textbf{H}(\textbf{p}) \geq \beta \}
\end{equation}
where $\alpha_e$ is the end-point search radius and $\beta$ is the heatmap threshold.
From the end-point sets, $\mathcal{E}_e(\textbf{e}^a_{1,2})$ and $\mathcal{E}_e(\textbf{e}^b_{1,2})$, sequential RANSAC \cite{fischler1981random} is employed to robustly detect two line segments parameterized by $\rho^e_{1,2}, \theta^e_{1,2}$.
The probabilistic observation model for time $t$ is defined in the line parameter space and written as a summation of Gaussians, similar to our previous work,
\begin{equation}
\begin{split}
    P(\rho^e_{1,2, t}, &\theta^e_{1,2, t} |  \textbf{b}_t, \textbf{w}_t) \propto \\ 
    & \sum_{i=1,2} e^{-\gamma_{\rho}|\rho^e_{i, t} - \rho_i(\textbf{b}_t, \textbf{w}_t)| - \gamma_\theta |\theta^e_{i, t} - \theta_i(\textbf{b}_t, \textbf{w}_t)| }
\end{split}
\end{equation}
where $\gamma_{\rho}$ and $\gamma_\theta$ describe the standard deviation of the $\rho^e_{1,2}$ and $\theta^e_{1,2}$ detections and $\rho_{1,2}(\textbf{b}_t, \textbf{w}_t), \theta_{1,2}(\textbf{b}_t, \textbf{w}_t)$ are the projected insertion-shaft lines from the estimated lumped error.
The projected insertion-shaft lines from the estimated lumped error come from (\ref{eq:fk}) and (\ref{eq:projected_cylinder_lines}) to generate line parameters $a_{1,2}(\textbf{b}_t, \textbf{w}_t), b_{1,2}(\textbf{b}_t, \textbf{w}_t), c_{1,2}(\textbf{b}_t, \textbf{w}_t)$ and subsequently converted as follows
\begin{equation}
\label{eq:line_parameter_prob}
\begin{split}
    \theta_{1,2}(\textbf{b}_t, \textbf{w}_t) &= - \tan^{-1} \left( \frac{a_{1,2}(\textbf{b}_t, \textbf{w}_t)}{b_{1,2}(\textbf{b}_t, \textbf{w}_t)} \right) \\
    \rho_{1,2}(\textbf{b}_t, \textbf{w}_t) &= - \frac{c_{1,2}(\textbf{b}_t, \textbf{w}_t)}{b_{1,2}(\textbf{b}_t, \textbf{w}_t)} \sin \left( \theta_{1,2}(\textbf{b}_t, \textbf{w}_t) \right) \; \; .
\end{split}
\end{equation}

\subsubsection{Line Intensities to Polar}
\label{sec:line_intensities_to_polar}

The Endpoint Intensities to Polar observation model is extended in this observation model by expanding the set of points to include points along the line segment rather than just near the detected endpoints to improve robustness by including more candidate points associated with the detected insertion-shaft line.
Written explicitly, the new set of points is defined as
\begin{equation}
    \label{eq:line_point_set}
	\mathcal{E}_l(\textbf{e}^a, \textbf{e}^b) = \{ \textbf{p} \hspace{1mm} \big| \hspace{1mm} \min_{l \in L(\textbf{e}^a, \textbf{e}^b)} || \textbf{p} - l || \leq \alpha_l \hspace{1mm} \cap \hspace{1mm} \textbf{H}(\textbf{p}) \geq \beta \}
\end{equation}
where $L(\textbf{e}^a, \textbf{e}^b)$ is the set of points on the line segment between $\textbf{e}^a, \textbf{e}^b$ and $\alpha_l$ is the search radius about the line.
Similar to the previous observation model, line segments parameters $\rho^l_{1,2}, \theta^l_{1,2}$ are derived from the line point sets, $\mathcal{E}_l(\textbf{e}^a_{1,2}, \textbf{e}^b_{1,2})$, using sequential RANSAC.
The same probability model in (\ref{eq:line_parameter_prob}) is used to define the observation model for the detected insertion-shaft lines, $\rho^l_{1,2,t}, \theta^l_{1,2,t}$, at timestep $t$.

\subsubsection{Endpoint Intensities}
\label{sec:endpoint_intensities}

The previous observation models use a noise distribution on derived line parameters.
Instead, the pixel based observation works directly in the pixel space by deriving a random variable $R(\cdot)$ which describes how well the detected pixel fits to the projected insertion-shaft,
\begin{equation}
    \label{eq:derived_random_variable}
    R( x,y | \theta, \rho) = \cos( \theta ) x + \sin( \theta ) y - \rho
\end{equation}
where $x,y$ is an input pixel coordinate.
The derived variable $R(\cdot)$ is inspired from \cite{chiu2022real} where a point is fitted to an ellipse projected from a suture needle.
The fitting-based derived random variables allow for a pixel-based probability distribution without requiring pixel-to-pixel association rather a pixel-to-line association.
We assume the noise of the pixels associated with the insertion-shaft's edges, $\textbf{p}$, are independent and normally distributed from the projected line, $\textbf{p} \sim \mathcal{N}([X,Y]^\top, \sigma^2)$ where $X,Y$ come from the projected insertion-shaft, i.e (\ref{eq:projected_cylinder_lines}).
Therefore, the derived random variable is Gaussian whose distribution is
\begin{equation}
    \label{eq:derived_random_variable_uncertainty}
    R(\textbf{p} | \theta, \rho ) \sim \mathcal{N} \left( 0, \cos( \theta )^2\sigma^2 + \sin( \theta )^2\sigma^2 \right)
\end{equation}
hence providing a probability distribution for a detected insertion-shaft pixel without requiring explicit pixel-to-pixel association.
This observation model applies the derived random variable over the endpoint sets of the detected insertion-shaft, $R(\textbf{p}_t | \theta_{1,2}(\textbf{b}_t, \textbf{w}_t), \rho_{1,2} (\textbf{b}_t, \textbf{w}_t)) \forall \textbf{p}_t \in \mathcal{E}_e(\textbf{e}^a_{1,2, t}), \mathcal{E}_e(\textbf{e}^b_{1,2, t})$, for a timestep $t$.

\subsubsection{Line Intensities}
\label{sec:line_intensities}

This observation model also uses the derived random variable $R(\cdot)$ defined in (\ref{eq:derived_random_variable}) with the same uncertainty distribution from (\ref{eq:derived_random_variable_uncertainty}).
The only difference with the previous observation model is the point sets which the derived random variable is applied to now covers the entire line, $R(\textbf{p}_t | \theta_{1,2}(\textbf{b}_t, \textbf{w}_t), \rho_{1,2} (\textbf{b}_t, \textbf{w}_t)) \; \forall \; \textbf{p}_t \in \mathcal{E}_l(\textbf{e}^a_{1,2, t}, \textbf{e}^b_{1,2, t})$, for a timestep $t$.

\begin{figure}
    \vspace{2mm}
    \centering 
    \begin{subfigure}{0.49\columnwidth}
            \includegraphics[width=\columnwidth, trim={0 0 11.5cm 0},clip]{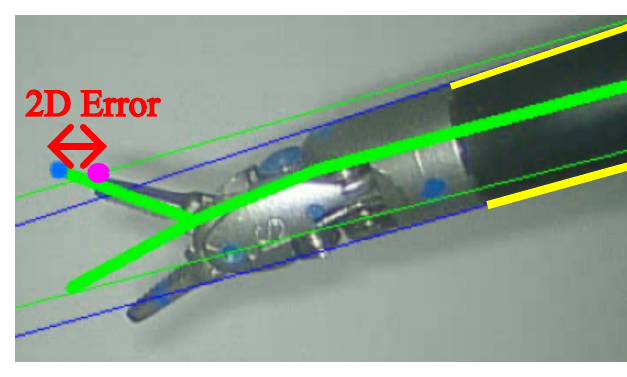}
    \end{subfigure}
    \begin{subfigure}{0.49\columnwidth}
            \includegraphics[width=\columnwidth, trim={11.5cm 0 0 0},clip]{dataset_images.pdf}
    \end{subfigure}
    \vspace{-3.5mm}
    \caption{Figures from our structured and deformable tissue datasets, from left to right respectively, where a green skeleton is overlaid to show the tracked surgical robotic tool in the scene and the yellow lines correspond to insertion-shaft line detections from our proposed approach.
    As shown in red, the 2D error metric for both datasets is calculated by comparing the projected and manually labelled inferior jaw tool tip.
    Note that the surgical tool in the deformable tissue dataset is intentionally far away to simulate a challenging surgical tool tracking scenario.}
    \label{fig:dataset_images}
\end{figure}


\begin{figure*}[t]
    \centering
    \vspace{2mm}
    \includegraphics[width=0.98\textwidth]{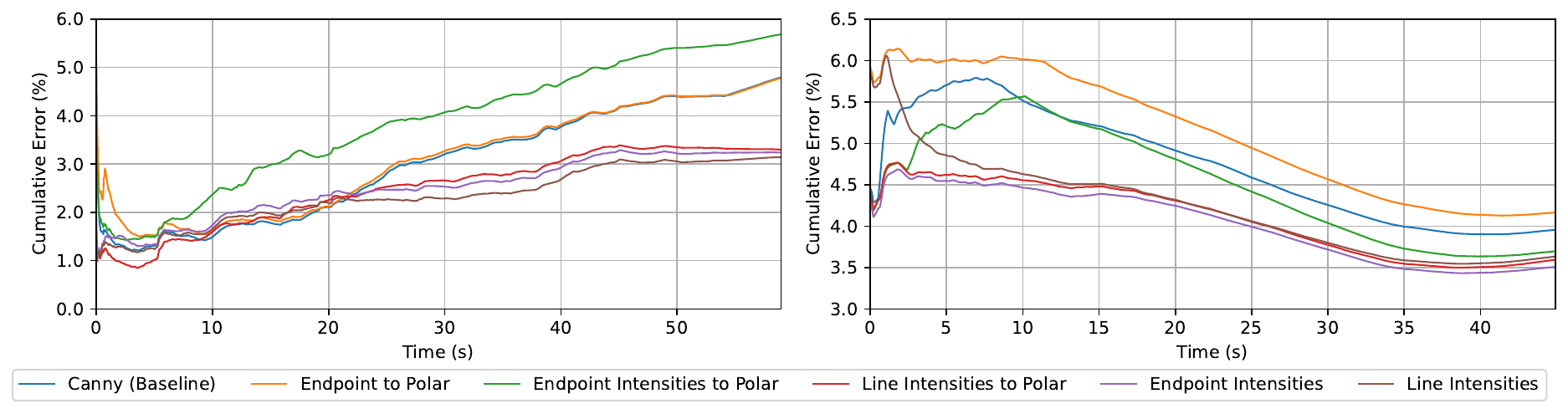}
    \caption{From left to right, the plots show the accumulated 2D error for Structured and Deformable Tissue Datasets, respectively. Our proposed line intensities to polar, endpoint intensities, and line intensity observation models are measured to provide consistent tracking results as seen in the plots. }
    \label{fig:2Daccumulated_error}
    \vspace{-2mm}
\end{figure*}

\begin{figure*}[t]
\centering
    \begin{subfigure}{0.32\textwidth}
        \includegraphics[width=1.0\textwidth, trim={0 1cm 0 1.5cm},clip]{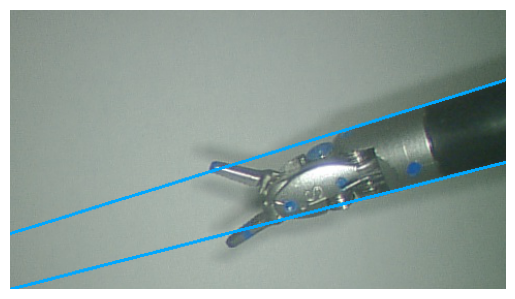}
        \caption{Canny (Baseline)}
        \vspace{1mm}
    \end{subfigure}
    \begin{subfigure}{0.32\textwidth}
            \includegraphics[width=1.0\textwidth, trim={0 1cm 0 1.5cm},clip]{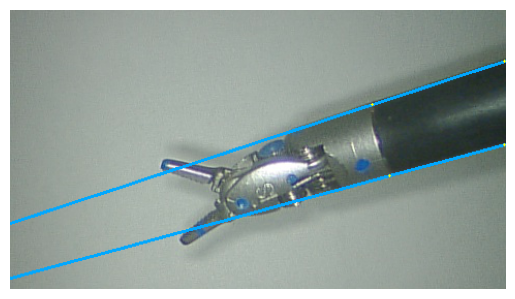}
        \caption{Endpoints to Polar}
        \vspace{1mm}
    \end{subfigure}
    \begin{subfigure}{0.32\textwidth}
            \includegraphics[width=1.0\textwidth, trim={0 1cm 0 1.5cm},clip]{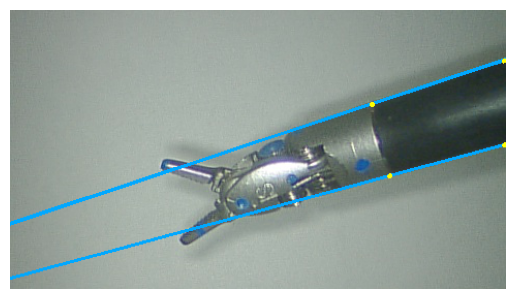}
        \caption{Endpoint Intensities to Polar}
        \vspace{1mm}
    \end{subfigure}
    \begin{subfigure}{0.32\textwidth}
        \includegraphics[width=1.0\textwidth, trim={0 1cm 0 1.5cm},clip]{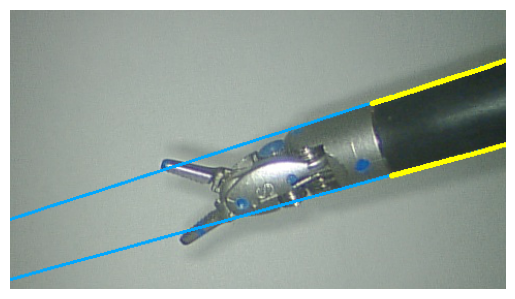}
        \caption{Line Intensities to Polar}
    \end{subfigure}
    \begin{subfigure}{0.32\textwidth}
            \includegraphics[width=1.0\textwidth, trim={0 1cm 0 1.5cm},clip]{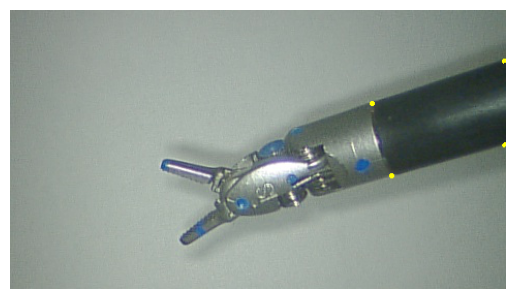}
        \caption{Endpoint Intensities}
    \end{subfigure}
    \begin{subfigure}{0.32\textwidth}
            \includegraphics[width=1.0\textwidth, trim={0 1cm 0 1.5cm},clip]{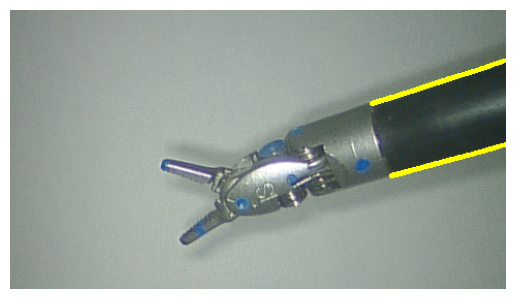}
        \caption{Line Intensities}
    \end{subfigure}
    \vspace{-1mm}
    \caption{Illustration of the different observation models for detecting the insertion shaft in the deformable tissue datasets.
    The blue lines highlight the parameterized lines (i.e. polar parameterization) and the yellow points represent the pixel set used in the observation model. 
    In such a structured scene, all insertion-shaft detection algorithms work well.
    This figure is best viewed in color.
    }
    \label{fig:algorithms_comp_1}
\end{figure*}

\section{EXPERIMENTS AND RESULTS}

A comparison experiment is ran to test the efficacy of the proposed methods by implementing the different observation models into a Particle Filter to track a surgical robotic tool.
The tracking algorithm compares across all the presented observation models in Sections \ref{sec:endpoint_intensities_to_polar} to \ref{sec:line_intensities} and the following baselines:
\begin{itemize}
    \item \textit{Canny}: previous surgical tool tracking approach \cite{richter2021robotic}
    \item \textit{Endpoint to Polar}: derived from Section \ref{sec:endpoint_intensities_to_polar} where $\alpha=0$ hence only relies on the endpoint detection algorithm from the DNN
\end{itemize}
In our implemented particle filter, we also include point feature detections replicated from our previously established approaches \cite{li2020super, richter2021robotic}.

\subsection{Implementation Details}

Motion model parameters are retained from our previous work \cite{richter2021robotic}.
The line parameter observation models used sequential RANSAC parameters of $5$ iterations, $3.0$ samples, a residual threshold of $0.75$, and maximum number of trials of $100$.
The pixel observation models used a radius of $\alpha = 10$ pixels and a heatmap threshold of $\beta = 0.90$.
The previous canny edge detection schema with hough accumulator used hyperparameters $\rho_{acc} = 5.0$, $\theta_{acc} = 0.09$, a vote threshold of $v = 100$, and cluster distances of $\rho_{cd} = 5.0$, $\theta_{cd} = 0.09$ as per \cite{richter2021robotic}. 

\begin{table}[t]
\vspace{-5mm}
    \caption{Surgical tool tracking results from structured environment and deformable tissue datasets.}
    \setlength\tabcolsep{0.5em}
    \centering
    \label{tab:results_table}
    \begin{tabular}{l  c  c  }
        \toprule
        \multicolumn{1}{c}{\multirow{2}{*}{\textbf{Method}}} & \textbf{Structured} & \textbf{Deformable Tissue} \\
         & 2D Error (\%) & 2D Error (\%)  \\
        \midrule
        \hspace{2mm} Canny Edge Detector \cite{richter2021robotic} &  4.8 (3.0) & 4.0 (1.2)  \\
        \hspace{2mm} Endpoint to Polar	& 4.8 (2.9) & 4.2 (1.3)  \\
        \hspace{2mm} Endpoint Intensities to Polar & 5.7 (2.9) & 3.7 (1.4) \\	
        \hspace{2mm} Line Intensities to Polar & 3.3 (2.0) & 3.6 (0.9)  \\	
        \hspace{2mm} Endpoint Intensities  & 3.2 (2.0) & 3.5 (0.9)  \\
        \hspace{2mm} Line Intensities & 3.1 (2.1) & 3.6 (0.9)  \\
        \bottomrule
    \end{tabular}
\end{table}

\begin{figure*}[t]
\centering
\vspace{2mm}
    \begin{subfigure}{0.32\textwidth}
        \includegraphics[width=1.0\textwidth, trim={0 1.5cm 0 0cm},clip]{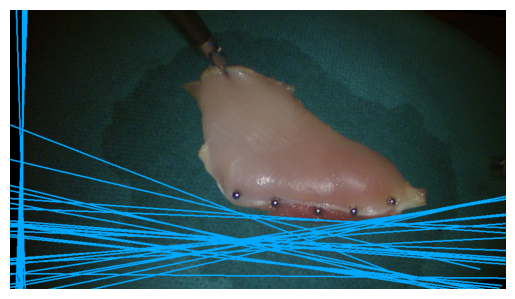}
        \caption{Canny (Baseline)}
    \end{subfigure}
    \begin{subfigure}{0.32\textwidth}
            \includegraphics[width=1.0\textwidth, trim={0 1.5cm 0 0cm},clip]{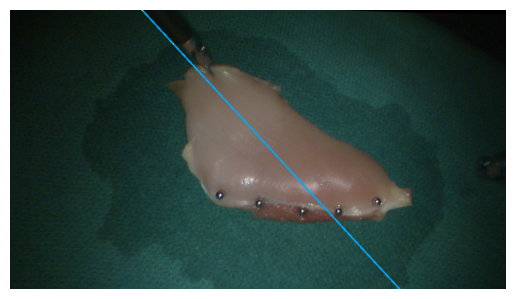}
        \caption{Endpoints to Polar}
    \end{subfigure}
    \begin{subfigure}{0.32\textwidth}
            \includegraphics[width=1.0\textwidth, trim={0 1.5cm 0 0cm},clip]{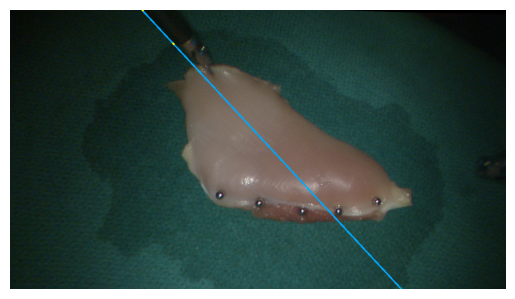}
        \caption{Endpoint Intensities to Polar}
    \end{subfigure}
    \begin{subfigure}{0.32\textwidth}
        \includegraphics[width=1.0\textwidth, trim={0 1.5cm 0 0cm},clip]{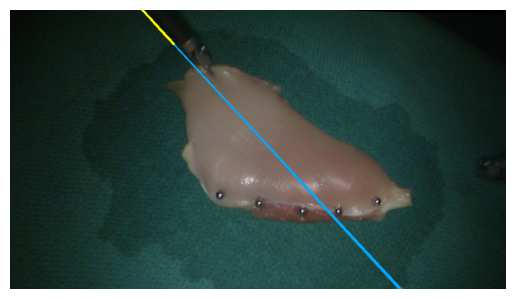}
        \caption{Line Intensities to Polar}
    \end{subfigure}
    \begin{subfigure}{0.32\textwidth}
            \includegraphics[width=1.0\textwidth, trim={0 1.5cm 0 0cm},clip]{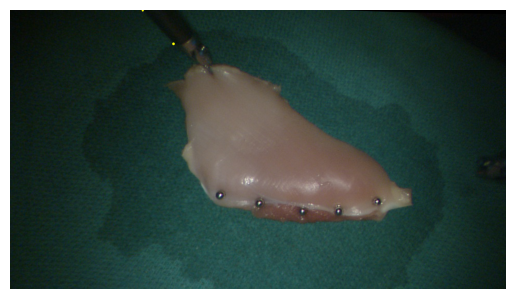}
        \caption{Endpoint Intensities}
    \end{subfigure}
    \begin{subfigure}{0.32\textwidth}
            \includegraphics[width=1.0\textwidth, trim={0 1.5cm 0 0cm},clip]{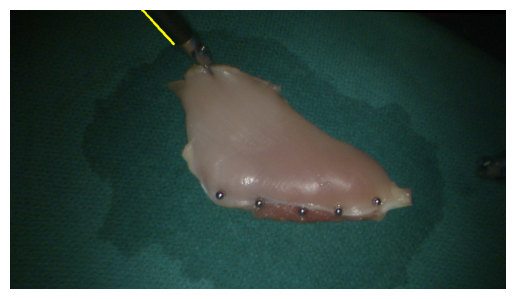}
        \caption{Line Intensities}
    \end{subfigure}
    \vspace{-1mm}
    \caption{Illustration of the different observation models for detecting the insertion shaft in the deformable tissue datasets. The blue lines highlight the parameterized lines (i.e. polar parameterization) and the yellow points represent the pixel set used in the observation model. 
    The surgical robotic tool in the deformable tissue dataset was intentionally made challenging to see to create a challenging low light scenario.
    Even in such a challenging scene, the proposed approach is still able to detect the insertion shaft.
    This figure is best viewed in color.
    }
    \vspace{-2mm}
    \label{fig:algorithms_comp_2}
\end{figure*}

\subsection{Datasets}

\subsubsection{Structured Environment Dataset}

To create our Structured Environment Dataset, we used a stationary dVRK \cite{kazanzides2014open} stereo endoscope to record a single PSM arm tool at various, teleoperated jaw and wrist configurations in $1080 \times 1920$ pixel resolution video at 30 frames-per-second (fps) over $60$ seconds. This dataset was collected for experiments in our prior work \cite{richter2021robotic} that established our benchmark tool tracking algorithm.
We chose to use a white background, a narrow angle field-of-view (FOV), and high intensity illumination to ensure adequate capture of surgical tool motion as well as tool tracking features. 
The pixel coordinates of the inferior jaw tool tip is labelled in each frame of the recorded video to create a ground truth reference for performance evaluation.
Performance is evaluated as a pixel distance error between the projected and manually labelled point as shown in Fig. \ref{fig:dataset_images}. Percentage error is calculated as the pixel distance error divided by the diagonal length of the image frame in which the pixel error was calculated.

\subsubsection{Deformable Tissue Dataset}

We used the same hardware and video specifications as noted in the Structured Environment Dataset.
The tool trajectory is from an autonomous trajectory for manipulating deformable soft tissue \cite{liu2021real}.
This scenario has lower intensity lighting, more surgical tool occlusions, and a wider FOV to mimic a more realistic surgical environment for automation.
The total length of this recording was $45$ seconds.
Similar to the Structured Environment Dataset, we manually labeled the pixel coordinates of the inferior jaw tool tip in each frame of the recorded video for a 2D error metric as shown in Fig. \ref{fig:dataset_images}. Pixel error and percentage error were calculated as in the Structured Dataset.

\subsection{Experimental Results}

Our Structured Environment Dataset results (Table 1) demonstrated that our Line Intensities algorithm had the lowest percentage error (3.1\%) relative to the other algorithms. The Endpoint Intensities algorithm had the second lowest percentage error (3.2\%) and the lowest standard error (2.0\%). With the exception of the Endpoint Intensities to Polar algorithm, all of our algorithms had lower error and standard error than the benchmark Canny algorithm.

We expected the benchmark canny edge detector algorithm to outperform given the extensive hyperparameter tuning for this specific dataset outlined in \cite{richter2021robotic}. However, only the Endpoint Intensities to Polar algorithm demonstrated worse performance than the benchmark on the Structured Dataset, likely due to the challenge of fitting multiple shaft lines to endpoint intensity points in a well-lit, high-quality image. In this environment, additional potential fitted lines from endpoints offer little information gain relative to the lines detected by the Canny benchmark.

Our Deformable Tissue Dataset results demonstrated that 2D error and standard error were lowest for our Endpoint Intensities algorithm: 3.5\% and 0.9\%, respectively. All of our algorithms, with the exception of Endpoint to Polar, had similar error performance and also outperformed the Canny benchmark. We expected the Endpoint to Polar algorithm to underperform in this environment given the difficulty in identifying endpoints in this more challenging environment. When an endpoint is not detected, the potential endpoint pair that defines a line is removed, and a potential shaft line is ignored. This decreased the performance of both endpoint-to-line fitting algorithms: Endpoint to Polar and Endpoint Intensities to Polar.
Endpoint Intensities, however, does not require a line-fitting approach so is able to compensate for missing shaft lines.


We also compute the accumulated error, which is normalized by total number of image data frames hence can decrease when performance improves, and this is shown in Fig. 4.
The accumulated error results for both datasets show that our algorithms outperform the benchmark in both 2D and 3D error metrics. These results are particularly important given the potential of tracking error to compound over time in autonomous surgery applications. In both datasets, our pixel-based methods had the lowest accumulated error. In the Structured Dataset, Line Intensities, Endpoint Intensities, and Line Intensities to Polar had 2D cumulative errors of 3.1\%, 3.3\%, and 3.2\%, respectively. In the Deformable Tissue Dataset, 2D cumulative errors for these algorithms were 3.6\%, 3.5\%, and 3.6\%, respectively.

\section{Discussion and Conclusion}

Our experimental results demonstrate that our tool tracking algorithms are not only more accurate in structured environments than the current state-of-the art approach but are also more generalizable to less structured environments.
The baseline approach, canny edge detection, has significant difficulty performing well in less structured environments such as low light scenarios.
Through our proposed insertion-shaft line detection approach, we are able to more robustly detect the insertion shaft in challenging scenarios as seen in Fig. \ref{fig:algorithms_comp_1} and \ref{fig:algorithms_comp_2}.
Furthermore, our proposed endpoint and line intensity based observation models are able to take full advantage of the predicted heatmap which why we believe they performed the best of all the comparisons.

Our previous tool tracking approach has already been used for significant surgical automation efforts including autonomous blood suction \cite{richter2021autonomous, huang2021model}, suture needle tracking \cite{chiu2022markerless, chiu2022real} and manipulation \cite{chiu2021bimanual}, real-to-sim registration \cite{liu2021real}, and more general surgical perception \cite{li2020super, lu2021super, lin2022semantic}.
For future work, we intend to continue these research efforts in less structured environments through our proposed novel surgical tool tracking approach. Limitations that will need to be addressed include inference and computation time as the SOLD2 line detection module performs at 1-2 fps.


\addtolength{\textheight}{0cm}   

\bibliographystyle{IEEEtran}
\bibliography{refs}

\end{document}